\documentclass{article}
\usepackage{spconf,amsmath,graphicx,times,CJK,multirow,color,pinyin,booktabs,latexsym}
\usepackage{ulem}
\usepackage{cite}
\usepackage{bbding}

\title{Modeling Homophone Noise for Robust Neural Machine Translation}
\name{
Wenjie Qin\textsuperscript{$\dagger$},
Xiang Li\textsuperscript{$\ddagger$}\sthanks{Xiang Li is the corresponding author and the work is done while Wenjie Qin is doing research intern at Xiaomi AI Lab}, 
Yuhui Sun\textsuperscript{$\ddagger$},
Deyi Xiong\textsuperscript{§}, 
Jianwei Cui\textsuperscript{$\ddagger$}, and
Bin Wang\textsuperscript{$\ddagger$}
}

\address{
\textsuperscript{$\dagger$}School of Computer Science and Technology, Soochow University, Suzhou, China \\
\textsuperscript{$\ddagger$}Xiaomi AI Lab, Beijing, China \\
\textsuperscript{§}College of Intelligence and Computing, Tianjin University, Tianjin, China \\
\tt wjqin@stu.suda.edu.cn\\
\tt $\lbrace$lixiang21, sunyuhui1, cuijianwei, wangbin11$\rbrace$@xiaomi.com \\
\tt dyxiong@tju.edu.cn
}

\begin{document}
%
\maketitle
\begin{abstract}
In this paper, we propose a robust neural machine translation (NMT) framework. The framework consists of a homophone noise detector and a syllable-aware NMT model to homophone errors. The detector identifies potential homophone errors in a textual sentence and converts them into syllables to form a mixed sequence that is then fed into the syllable-aware NMT. Extensive experiments on Chinese$\rightarrow$English translation demonstrate that our proposed method not only significantly outperforms baselines on noisy test sets with homophone noise, but also achieves a substantial improvement on clean text.
\end{abstract}
\section{Introduction}
\label{sec:intro}

Despite of remarkable progress made in NMT recently ~\cite{ bahdanau2015neural,DBLP:conf/emnlp/LuongPM15, DBLP:conf/icml/GehringAGYD17,DBLP:conf/nips/VaswaniSPUJGKP17,edunov-etal-2018-understanding,zhang2019future}, most NMT systems are still prone to translation errors caused by noisy input sequences~\cite{DBLP:conf/iclr/BelinkovB18,DBLP:conf/aclnmt/KhayrallahK18}. One common type of input noise is homophone noise, where words or characters are mis-recognized as others with same or similar pronunciation in ASR~\cite{goldwater2010words,rohit2012active} or input systems~\cite{zhang-etal-2019-open} for non-phonetic languages (e.g., Chinese), as illustrated by the example in Table~\ref{table:NMT defeated case}. 

Previous works suggest that incorporating phonetic embeddings into NMT~\cite{liu2019robust} and augmenting training data with adversarial examples~\cite{DBLP:conf/iclr/BelinkovB18} with injected homophone noise~\cite{li2018improve} would alleviate this issue. Intuitively, humans usually have no trouble in disambiguating sentences corrupted with moderate homophone noise via context and syllable information. We propose a human-inspired robust NMT framework tailored to homophone noise for Chinese-English translation, which is composed of a homophone noise detector (hereinafter referred to as detector for brevity) and a syllable-aware NMT (SANMT) model.

\begin{table}[]
\centering
\begin{tabular}{ll}
\toprule[1pt]
Clean Input~&~\begin{CJK}{UTF8}{gbsn}建一所小学 \end{CJK}   \\
Output of NMT~&~build a primary school \\ \specialrule{0.05em}{3pt}{3pt}
Noisy Input~&~\begin{CJK}{UTF8}{gbsn}建\uline{议}所小学 \end{CJK} \\ 
Output of NMT~&~suggest a primary school \\ 
\specialrule{0.05em}{3pt}{3pt}
Mixed Transcript~&~\begin{CJK}{UTF8}{gbsn}建\uline{yi}所小学 \end{CJK} \\ 
Output of Ours~&~build a primary school\\
\bottomrule[1pt]
\end{tabular}
\caption{A translation example with homophone noise. The erroneous character ``\begin{CJK}{UTF8}{gbsn}议\end{CJK}'' (``discuss'') in the noisy input is a homophone corresponding to the original character ``\begin{CJK}{UTF8}{gbsn}一\end{CJK}'' (``one'') in the clean input. The erroneous character ``\begin{CJK}{UTF8}{gbsn}议\end{CJK}'' is replaced with its Chinese Pinyin ``yi'' in the mixed transcript which enables NMT to translate correctly.} \label{table:NMT defeated case}
\end{table}

Due to the lack of data annotated with homophone noise, we propose to train our detector on monolingual data in a self-supervised manner, where Chinese characters sequences as input and their corresponding syllables sequence as label to predict the possibility that a character is homophone noise. The identified homophone errors from a source sentence are then converted into corresponding syllables to produce a new source sequence mixed with characters and syllables. Augmenting bilingual training data with instances where original source sentences are substituted with their corresponding character-syllable-mixed sequences, we train the SANMT model to translate such unconventional inputs. To examine the effectiveness of our proposed model, we conduct extensive experiments on both artificial noisy test sets and a real-world noise test set with homophone noise in speech translation (ST) scenario. The test set will be released soon. Our experimental results on Chinese$\rightarrow$English translation clearly show that the proposed method is not only significantly superior to previous approaches in alleviating the impact of homophone noise on NMT, but also achieves a substantial improvement on the clean text.

\begin{figure}[h]
	\centering
 	\scalebox{0.33}{\includegraphics{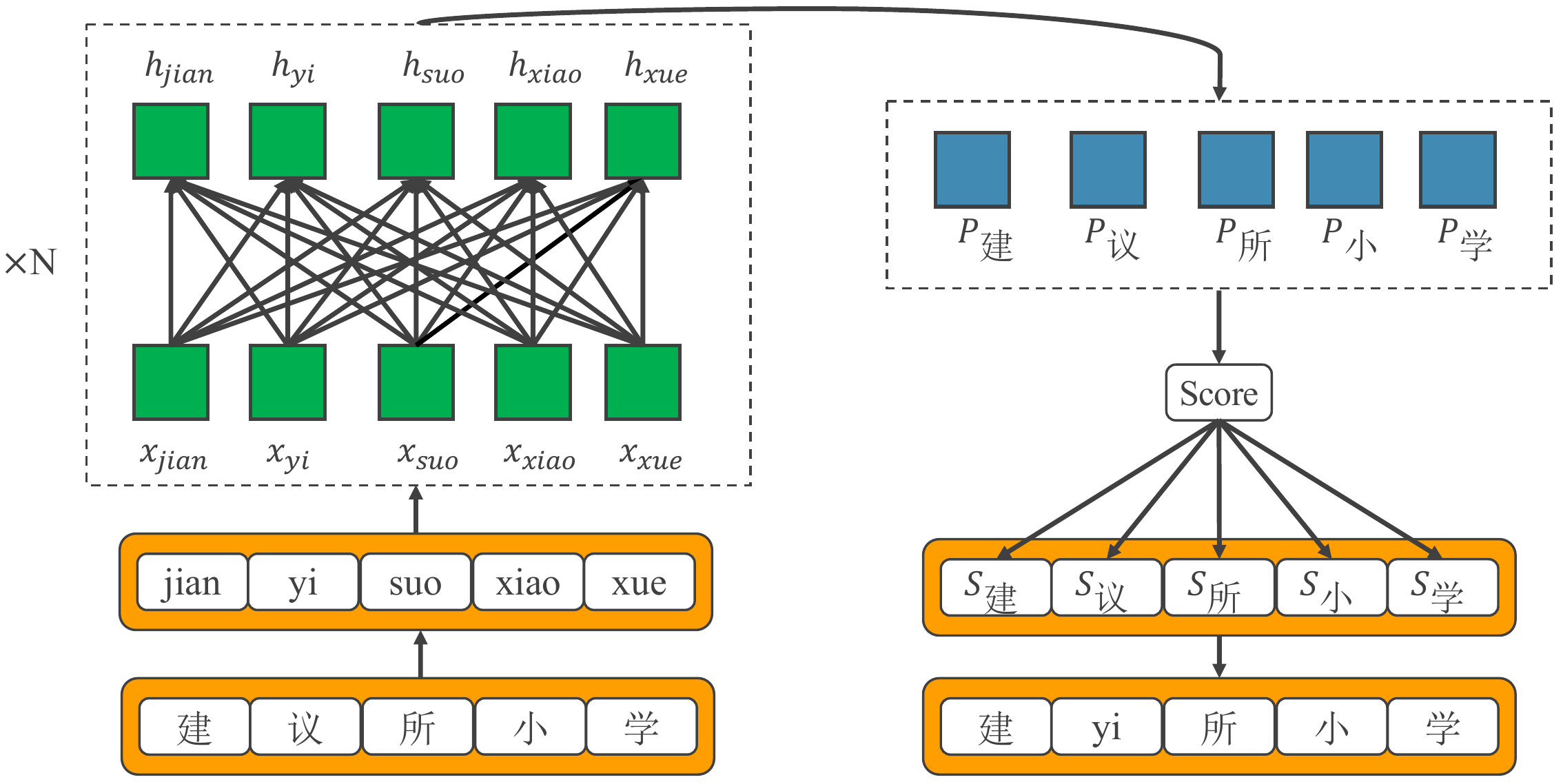}}
	\caption{The diagram of our proposed detector. Each character's log-likelihood score (LLS) is computed over the whole syllable sequence. In this example, ``\begin{CJK}{UTF8}{gbsn}议\end{CJK}'' is identified as a homophone error and replaced with its syllable ``\textit{yi}'', resulting in a new mixed sentence. }
	\label{detector} 
\end{figure}

\section{Approach}
\label{sec:approach}

\subsection{Homophone Error Detector}

Instead of feeding the noise input indiscriminately to NMT, we actively detect homophone errors in the inputs and roll back identified erroneous characters into syllables (i.e., Chinese Pinyin which consists of 400 syllables without tones representing about 6,500 characters) to minimise the interference of these non-negligible textual errors with NMT. In order to avoid the negative impact from textual noise on the detection, we predict each character $x_i$ in an original textual sentence $X=(x_1,x_2,...,x_T)$ based on the corresponding clean syllable sequence $S=(s_1,s_2,...,s_T)$ with the same length of $X$, instead of using $X$. Therefore, we first build a syllable-character parallel training data by converting the text-only corpus into a syllable corpus using a language-specific character-to-syllable converter. In this work, we utilize the Transformer~\cite{DBLP:conf/nips/VaswaniSPUJGKP17} encoder to model the interactions of syllables for our detector. Figure~\ref{detector} shows the detector architecture. The detector is trained using the standard maximum-likelihood criterion, to maximize the sum of log probabilities of the ground-truth outputs given the corresponding syllable inputs.

During inference, we use the LLS of each syllable token based on the whole syllable sequence as a confidence score that determines if the corresponding original textual character is a homophone error as follows:
\begin{equation}
LLS(x_{i})=logP(x_{i}|S;\theta),~
\end{equation}
where $\theta$ represents the parameter set of the detector. For a source sentence probably with homophone noise, we substitute the characters with LLS smaller than a predefined threshold $\beta$ with their corresponding syllables, forming a new sequence mixed with characters and syllables to downstream NMT.

\subsection{Syllable-aware NMT}

The previous works~\cite{li2018improve, liu2019robust} on robust NMT to homophone noise create the noise adversarial training data by injecting some homophone noise into the source part of original textual training data. Their resulting NMT models are able to handle inputs with homophone errors innately. However, our proposed detector is expected to convert identified erroneous characters in such inputs into corresponding syllables. Thus, such mixed inputs are not supported by conventional NMT models solely trained on the original textual training data or noise adversarial training data. To deal with this issue, we propose a syllable-aware NMT based on the Transformer~\cite{DBLP:conf/nips/VaswaniSPUJGKP17} encoder-decoder architecture to handle both conventional textual inputs and unconventional sequences mixed with syllables generated by the detector. Without introducing any changes either to the architecture or to the training scheme, we confer robustness on our SANMT model through augmenting the original training data with syllable-mixed training data. Specifically, we pick sentence pairs from original training data, and change the source sentences by randomly substituting some characters with corresponding syllables to create our syllable-mixed training data. It is worth noting that injected consecutive syllables are separated by blank space. Examples can be seen for each kind of training data in Table~\ref{table:data case}. 

\begin{table}[]
\small
\centering
\setlength{\tabcolsep}{1.0mm}{
\begin{tabular}{l|c|c}
\toprule[1pt]
\textbf{Data}    ~&~ \textbf{Source} ~&~ \textbf{Target} \\ \hline
OTTD    ~&~ \begin{CJK}{UTF8}{gbsn}建一所小学\end{CJK} ~&~ build a primary school \\
SMTD    ~&~ \begin{CJK}{UTF8}{gbsn}建 \uline{yi} 所小学\end{CJK} ~&~ build a primary school  \\
NATD    ~&~ \begin{CJK}{UTF8}{gbsn}建\uline{议}所小学\end{CJK} ~&~ build a primary school \\
\bottomrule[1pt]
\end{tabular}
}
\caption{Examples of original textual training data (OTTD), syllable-mixed training data (SMTD) and noise adversarial training data (NATD). The dashed underlined syllable ``\textit{yi}'' and character ``\begin{CJK}{UTF8}{gbsn}议\end{CJK}'' are the injected noise to substitute the original clean character ``\begin{CJK}{UTF8}{gbsn}一\end{CJK}'' in the source part of OTTD.}
\label{table:data case}
\end{table}

\begin{table*}[]
\small
\centering
\setlength{\tabcolsep}{4mm}
\resizebox{\textwidth}{!}{
\begin{tabular}{ll|c|cccccc}
\toprule[1pt]
\textbf{Model}                                                                                        &    & \textbf{MT02} & \textbf{MT03} & \textbf{MT04} & \textbf{MT05} & \textbf{MT06} & \textbf{MT08} & \textbf{Avg.}   \\ \hline
 
\multirow{2}{*}{Baseline}                                                                    & Clean & 47.23  & 46.82  & 47.69  & 45.81  & 44.84  & 35.82  & 44.20 \\ 
          & Noisy  & 34.84  & 34.18  & 35.35  & 32.31  & 32.39  & 26.63  & 32.17 \\ \hline
\multirow{2}{*}{CPNMT}                                                                    & Clean & 45.25  & 44.14  & 45.84  & 42.56  & 41.81  & 32.30  & 41.33 \\ 
& Noisy & 45.25  & 44.14  & 45.84  & 42.56  & 41.81  & 32.30  & 41.33 \\ \hline        
\multirow{2}{*}{Li et al \cite{li2018improve}}                                                       & Clean & 46.58  & 46.09  & 47.46  & 45.48  & 44.68  & 35.81  & 43.90 \\ 
         & Noisy  & 46.27  & 45.89  & 46.96  & 44.84  & 44.16  & 34.73  & 43.32 \\ \hline
\multirow{2}{*}{Liu et al \cite{liu2019robust}}                                                      & Clean & 46.24  & 44.67  & 46.80  & 44.08  & 43.00  & 34.12  & 42.53 \\ 
         & Noisy  & 46.00  & 44.79  & 46.55  & 43.66  & 42.80  & 33.88  & 42.34 \\ \hline
\multirow{2}{*}{Ours}                                                                       & Clean & 48.34  & 47.93  & 48.67  & 46.42  & 46.39  & 37.65  & 45.41 \\ 
         & Noisy  & \textbf{48.12}  & \textbf{46.65}  & \textbf{48.01}  & \textbf{45.61}  & \textbf{46.07}  & \textbf{36.57}  & \textbf{44.58 } \\
\bottomrule[1pt]
\end{tabular}}
\caption{Results of different NMT systems trained on LDC. Bold indicates the highest BLEU scores on ANTs.}
\label{table:subword-level}
\end{table*}

\section{Experiments}
\label{sec:experiments}

Despite its rareness in English and other Indo-European languages, homophone is a common phenomenon in Chinese. In this section, we conducted experiments on Chinese$\rightarrow$English translation to evaluate the effectiveness of our proposed method.

\subsection{Setup}

We used two training data sets for NMT, including (1) 1.2M sentence pairs extracted from LDC corpora and (2) CWMT corpus\footnote{http://www.statmt.org/wmt19/translation-task.html} consisting of 9M sentence pairs. We chose the NIST 2002 data set as our validation set, and NIST 2003, 2004, 2005, 2006, and 2008 data sets as our clean test sets (CTs). We built test sets with artificial noise (ANTs) by randomly substituting a small fraction of each source sentence in CTs with erroneous homophones. According to the experimental results reported in~\cite{park2019specaugment} that the state-of-the-art ASR performance is about 6.8\%-14.6\% word error rate on different test sets, we set 0.1 as the default noise ratio in our experiments to simulate as closely as possible the conditions of imperfect ASR. We used a Chinese Pinyin tool\footnote{https://github.com/mozillazg/python-pinyin} to convert Chinese characters into syllables.

We applied the standard Transformer-base architecture~\cite{DBLP:conf/nips/VaswaniSPUJGKP17} with its default hyper-parameters for NMT. We used the tokenization script provided in Moses\footnote{http://www.statmt.org/moses/} and Jieba tokenizer\footnote{https://github.com/fxsjy/jieba} for English and Chinese tokenizations, respectively. We learned BPE~\cite{DBLP:conf/acl/SennrichHB16a} models with 30K merge operations for both sides. We batched sentence pairs by approximately the same length and limited the number of source and target tokens per batch to 4,096. We set a beam size of 4 for decoding. We used case-insensitive tokenized BLEU~\cite{papineni-etal-2002-bleu} score as metric.

We implemented our detector based on the standard BERT-base model~\cite{DBLP:conf/naacl/DevlinCLT19}. The Chinese part of CWMT corpus is used as text-only data to train the detector. We set  $\beta$ to 0.1 as the threshold for the detector to identify the potential homophone errors.

\subsection{Robustness to Artificial Noise}

We used three times the original textual training data to build the syllable-mixed training data, resulting in an augmented training data 4-fold for training our SANMT. In detail, each source sentence of the syllable-mixed training data is constructed by randomly replacing some characters of its original one with their syllables according to a ratio $p\in(0,1)$. For a fair comparison, we created the noise adversarial training data based on the syllable-mixed training data combining with original training data to train the previous robust NMT models~\cite{li2018improve, liu2019robust}. When constructing noise adversarial training data, for each injected syllable in the syllable-mixed training data, we randomly replaced an original character with a different homophonic character. Following to~\cite{liu2019robust}, we set the weights of phonetic embedding and textual embedding to 0.95 and 0.05, respectively.

Table~\ref{table:subword-level} shows the results. Compared with the performance on CTs, the conventional NMT baseline suffers a huge drop on ANTs by up to 12 BLEU points. Among all methods, our approach not only achieves the best result across ANTs, especially up to an average improvement of 12.41 BLEU points compared to the NMT baseline, but also obtains a substantial improvement across CTs.  Although~\cite{liu2019robust} also achieve a considerable improvement on ANTs, it is a little worse than the simple data augmentation method~\cite{li2018improve} which doesn't introduce additional phonetic embeddings.

For a noisy input in which all the characters are homophone errors, the input can be completely rolled back a sequence only with syllables using our detector. Thus, we investigated the worst performance of our SANMT by training another NMT model using Pinyin as the modeling unit, namely character-level Pinyin NMT (CPNMT). Although CPNMT performs the worst among all the systems on CTs, it also significantly outperforms the baseline on ANTs. However, it still lags behind all the robust NMT systems. This demonstrates that both textual representation and syllable representation play an important role in improving translation quality and enhancing robustness, respectively. We also reproduced the experiment on the CWMT corpus and Table ~\ref{table:subword-level-CWMT} shows the main results. 

\begin{table*}[]
\small
\centering
\setlength{\tabcolsep}{4mm}
\resizebox{\textwidth}{!}{
\begin{tabular}{ll|c|ccc}
\toprule[1pt]
\textbf{Model}                                                                                        &    & \textbf{newstest17} & \textbf{newstest18} & \textbf{newstest19} & \textbf{Avg.}   \\ \hline
 
\multirow{2}{*}{Baseline}                                                                    & Clean & 22.59  & 21.95  & 24.85   & 23.40 \\ 
          & Noisy  & 6.60  & 6.28  & 6.03  & 6.16 \\ \hline
\multirow{2}{*}{Li et al \cite{li2018improve}}                                                       & Clean & 22.14  & 21.78  & 24.01  & 22.90 \\ 
         & Noisy  & 16.51  & 16.56  & 17.48  & 17.02 \\ \hline
\multirow{2}{*}{Ours}                                                                       & Clean & 22.37  & 21.84  & 24.24  & 23.04 \\ 
         & Noisy  & \textbf{20.72}  & \textbf{20.15}  & \textbf{22.51}  & \textbf{21.33} \\
\bottomrule[1pt]
\end{tabular}}
\caption{Results of different NMT systems trained on CWMT. Bold indicates the highest BLEU scores on ANTs.}
\label{table:subword-level-CWMT}
\end{table*}

\subsection{Effect of Noise Ratio}

We reported the averaged BLEU score of all systems on ANTs when varying the noise ratio from 0.1 to 0.5 in Table~\ref{table:noise fraction}. The results indicate that our method still performs the best at all settings. Not surprisingly, the baseline NMT system degrades more severely when the noise fraction becomes larger, but all robust NMT systems maintains relatively stable performance. 

\begin{table}[]
\centering
\setlength{\tabcolsep}{2.0mm}
\begin{tabular}{l|c|c|c|c|c}
\toprule[1pt]
\textbf{Model}    & 0.1            & 0.2   & 0.3   & 0.4   & 0.5   \\ \hline
Baseline & 32.17 & 22.00 & 14.86 & 10.19 & 7.09  \\ \hline
Li et al \cite{li2018improve}       & 43.32          & 42.96 & 42.51 & 42.07 & 41.62 \\ \hline
Liu et al \cite{liu2019robust}     & 42.34          & 41.97 & 41.92 & 41.52 & 41.38 \\ \hline
Our SANMT     & \textbf{44.58}          & \textbf{44.30} & \textbf{44.34} & \textbf{43.05} & \textbf{42.34} \\
\bottomrule[1pt]
\end{tabular}
\caption{The averaged BLEU scores of different NMT systems on ANTs with different noise ratios.}
\label{table:noise fraction}
\end{table}

\begin{table}[]
\small
\centering
\setlength{\tabcolsep}{2.0mm}
\begin{tabular}{l|c|c}
\toprule[1pt]
\textbf{Model}    ~~&~~ \textbf{Training Data} ~&~ \textbf{RNT} \\ \hline
\multirow{2}{*}{Baseline} ~~&~~ LDC        ~&~ 18.85                 \\ 
                          ~~&~~ CWMT        ~&~ 28.56                 \\ \hline
\multirow{2}{*}{Li et al \cite{li2018improve}}   ~~&~~ LDC       ~&~ 23.20                 \\ 
                          ~~&~~ CWMT        ~&~ 37.13                 \\ \hline
\multirow{2}{*}{Liu et al \cite{liu2019robust}}   ~~&~~ LDC       ~&~ 22.13                \\ 
                          ~~&~~ CWMT        ~&~ 38.12                 \\ \hline
\multirow{2}{*}{Our SANMT}    ~~&~~ LDC        ~&~ \textbf{25.36}                 \\
                          ~~&~~ CWMT        ~&~ \textbf{40.89}                 \\
\bottomrule[1pt]
\end{tabular}
\caption{Results of different NMT systems on RNT.}
\label{table:real scenario}
\end{table}

\subsection{Robustness to Real-World Noise}

There are no publicly available dedicated noisy test sets to examine the robustness of NMT to homophone errors. And the distribution of homophone noise in real-world scenarios is usually different from that in our simulated ANTs. Therefore, we propose a benchmark data set, namely test set with real-world noise (RNT) to facilitate research in this area. We manually collected noisy transcriptions with at least one homophone error recognized by human from our in-house Chinese ASR system, and commissioned a professional translator to translate them into English, resulting in RNT with 850 test samples. To further validate the effectiveness of different NMT systems when larger training data is available, we additionally trained all NMT models on the CWMT corpus. The results shown in Table~\ref{table:real scenario} clearly show that our method still significantly outperforms baselines on RNT.

\begin{table}[]
\small
\centering
\setlength{\tabcolsep}{0.6mm}
\begin{tabular}{p{3cm}p{4.5cm}}
\toprule[1pt]
Clean Utterance & \begin{CJK}{UTF8}{gbsn}请拼写它\end{CJK} \\
Noisy Transcript & \begin{CJK}{UTF8}{gbsn}请拼写\uline{他}\end{CJK} \\           Li et al \cite{li2018improve} & Please spell him   \\
Liu et al \cite{liu2019robust} & Please spell him   \\
Output of Detector & \begin{CJK}{UTF8}{gbsn}请 pin xie ta\end{CJK}  \\
Output of SANMT & Please spell it  \\ \specialrule{0.05em}{3pt}{3pt}
Clean Utterance  & \begin{CJK}{UTF8}{gbsn}听医生的建议\end{CJK} \\
Noisy Transcript & \begin{CJK}{UTF8}{gbsn}听\uline{一}生的建议\end{CJK} \\
Li et al \cite{li2018improve} & Listen to the doctor's advice  \\ 
Liu et al \cite{liu2019robust} & Suggestions on life  \\ 
Output of Detector & \begin{CJK}{UTF8}{gbsn}听 yi 生的建议\end{CJK} \\
Output of SANMT & Listen to doctors' advice \\
\bottomrule[1pt]
\end{tabular}
\caption{Cases with different types of homophone errors in RNT and their translations. The upper case denotes a third-person pronoun error, which is common in Chinese ASR where ``\begin{CJK}{UTF8}{gbsn}他\end{CJK}'', ``\begin{CJK}{UTF8}{gbsn}她\end{CJK}'', and ``\begin{CJK}{UTF8}{gbsn}它\end{CJK}'' are all pronounced ``t\=a''. The erroneous homophone word ``\begin{CJK}{UTF8}{gbsn}一生\end{CJK}'' in the lower case is also a noun which has a different meaning from the original homophone one ``\begin{CJK}{UTF8}{gbsn}医生\end{CJK}''.}
\label{table:case study}
\end{table}

Table~\ref{table:case study} shows two representative examples with different types of homophone errors from RNT. Results indicates that our detector is able to distinguish homophone noise and our SANMT generates correct translations using the syllable-mixed inputs. Thus, our robust NMT is superior to previous work in the real-world scenarios. 

\section{Conclusion}
\label{sec:conclusion}

In this paper, we have presented a novel framework composed of a homophone error detector and a SANMT model to cope with homophone noise. Experimental results show that our method not only achieves substantial improvement over previous robust NMT baselines both on the test sets with artificial or real-world noise, but also outperforms the NMT baseline on the clean test sets. We consider that future studies could modeling noise detection and NMT jointly.

\bibliographystyle{IEEEbib}
\bibliography{ICASSP2021}
\end{document}